%% file: main.tex
\documentclass[letterpaper]{article} 
\usepackage{aaai24}  
\usepackage{times}  
\usepackage{helvet}  
\usepackage{courier}  
\usepackage[hyphens]{url}  
\usepackage{graphicx} 
\usepackage{mathtools} 
\usepackage{amsfonts}  
\usepackage{xspace}
\usepackage{subfig}
\usepackage{multirow}
\usepackage{color}
\usepackage{xcolor}
\usepackage[T1]{fontenc}

\usepackage{natbib}  
\usepackage{caption} 
\usepackage{algorithm}
\usepackage{algorithmic}

\usepackage{newfloat}
\usepackage{listings}
\DeclareCaptionStyle{ruled}{labelfont=normalfont,labelsep=colon,strut=off} 
\floatstyle{ruled}
\newfloat{listing}{tb}{lst}{}
\floatname{listing}{Listing}
\title{Unveiling Invariances via Neural Network Pruning}
\author {
    Derek Xu\textsuperscript{\rm 1},
    Yizhou Sun\textsuperscript{\rm 1},
    Wei Wang\textsuperscript{\rm 1}
}
\affiliations {
    \textsuperscript{\rm 1}University of California, Los Angeles\\
    derekqxu@cs.ucla.edu,
    yzsun@cs.ucla.edu,
    weiwang@cs.ucla.edu
}

\usepackage{bibentry}

\def\SPSB#1#2{\rlap{\textsuperscript{#1}}\SB{#2}}
\def\SP#1{\textsuperscript{#1}}
\def\SB#1{\textsubscript{#1}}

\newcommand{\our}{\textsc{IUNet}\xspace}
\newcommand{\ilo}{\textsc{ILO}\xspace}
\newcommand{\pis}{\textsc{PIs}\xspace}
\newcommand{\betalasso}{$\beta$-\textsc{Lasso}\xspace}
\newcommand{\omp}{\textsc{OMP}\xspace}

\newcommand{\xgb}{\textsc{XGB}\xspace}

\newcommand{\tabn}{\textsc{TabN}\xspace}

\newcommand{\mlpv}{\textsc{MLP}\SB{\textsc{vis}}\xspace}
\newcommand{\ourmlpv}{\our\SP{(\mlpv)}\xspace}
\newcommand{\betalassomlpv}{\betalasso\SP{(\mlpv)}\xspace}
\newcommand{\ompmlpv}{\omp\SP{(\mlpv)}\xspace}
\newcommand{\ourmlpvP}{\our\SPSB{(\mlpv)}{\textsc{no-prune}}\xspace}
\newcommand{\ourmlpvILO}{\our\SPSB{(\mlpv)}{\textsc{no-\ilo}}\xspace}
\newcommand{\ourmlpvPIS}{\our\SPSB{(\mlpv)}{\textsc{no-\pis}}\xspace}

\newcommand{\resnet}{\textsc{ResNet}\xspace}
\newcommand{\ourres}{\our\SP{(\resnet)}\xspace}
\newcommand{\betalassores}{\betalasso\SP{(\resnet)}\xspace}
\newcommand{\ompres}{\omp\SP{(\resnet)}\xspace}

\newcommand{\mlpt}{\textsc{MLP}\SB{\textsc{TAB}}\xspace}
\newcommand{\ourmlpt}{\our\SP{(\mlpt)}\xspace}
\newcommand{\betalassomlpt}{\betalasso\SP{(\mlpt)}\xspace}
\newcommand{\ompmlpt}{\omp\SP{(\mlpt)}\xspace}
\newcommand{\mlpc}{\textsc{MLP}\SB{\textsc{tab}+C}\xspace}
\newcommand{\ourP}{\our\SB{\textsc{no-prune}}\xspace}
\newcommand{\ourmlptP}{\our\SPSB{(\mlpt)}{\textsc{no-prune}}\xspace}
\newcommand{\ourmlptILO}{\our\SPSB{(\mlpt)}{\textsc{no-\ilo}}\xspace}
\newcommand{\ourmlptPIS}{\our\SPSB{(\mlpt)}{\textsc{no-\pis}}\xspace}

\begin{document}

\maketitle

\begin{abstract}
Invariance describes transformations that do not alter data's underlying semantics. Neural networks that preserve natural invariance capture good inductive biases and achieve superior performance. Hence, modern networks are handcrafted to handle well-known invariances (ex. translations). We propose a framework to learn novel network architectures that capture data-dependent invariances via pruning. Our learned architectures consistently outperform dense neural networks on both vision and tabular datasets in both efficiency and effectiveness. We demonstrate our framework on multiple deep learning models across 3 vision and 40 tabular datasets.
\end{abstract}

\input{sec-introduction.tex}

\input{sec-related.tex}

\input{sec-model.tex}

\input{sec-exp.tex}

\input{sec-result.tex}

\input{sec-conclusion.tex}

\input{sec-supp-related.tex}
\input{sec-supp-loss.tex}
\input{sec-supp-exp.tex}
\input{sec-supp-implementation.tex}

\bibliography{aaai24}

\end{document}

%% file: sec-introduction.tex
\section{Introduction} 


Preserving invariance is a key property in successful neural network architectures. Invariance occurs when the semantics of data remains unchanged under a set of transformations~\citep{bronstein2017geometric}. For example, an image of a cat can be translated, rotated, and scaled, without altering its underlying contents. Neural network architectures that represent data passed through invariant transformations with the same representation inherit a good inductive bias~\citep{neyshabur2020towards,neyshabur2017implicit,neyshabur2014search} and achieve superior performance~\citep{zhang2021understanding,arpit2017closer}.

Convolutional Neural Networks (CNNs) are one such example. CNNs achieve translation invariance by operating on local patches of data and weight sharing. Hence, early CNNs outperform large multilayer perceptrons (MLP) in computer vision~\citep{lecun2015deep,lecun1998gradient}. Recent computer vision works explore more general spatial invariances, such as rotation and scaling~\citep{satorras2021n,deng2021vector,delchevalerie2021achieving,sabour2017dynamic,cohen2016group,jaderberg2015spatial, qi2017pointnet++,jaderberg2015spatial,xu2014scale}. Other geometric deep learning works extend CNNs  to non-Euclidean data by considering additional data-type specific invariances, such as permutation invariance~\citep{wu2020comprehensive,kipf2016semi,defferrard2016convolutional}.



Designing invariant neural networks requires substantial human effort: both to determine the set of invariant transformations and to handcraft architectures that preserve said transformations. In addition to being labor-intensive, this approach has not yet succeeded for all data-types~\citep{schafl2022hopular,gorishniy2022embeddings,gorishniy2021revisiting,huang2020tabtransformer}. For example, designing neural architectures for tabular data is especially hard because the set of invariant tabular transformations is not clearly-defined. Thus, the state-of-the-art deep learning architecture on tabular data remains MLP~\citep{kadra2021well, grinsztajn2022tree}.

Existing invariance learning methods operate at the data augmentation level~\citep{immer2022invariance,quiroga2020revisiting,benton2020learning,cubuk2018autoaugment}, where a model is trained on sets of transformed samples rather than individual samples. This makes the network resiliant to invariant transformations at test time. Contrastive learning (CL) is shown to be an effective means of incorporating invariance~\citep{dangovski2021equivariant}, and has seen success across various tasks~\citep{chen2021empirical,zhu2021graph,you2020graph,jaiswal2020survey,baevski2020wav2vec,chen2020simple}, including tabular learning~\citep{bahri2021scarf}. While these approaches train existing neural networks to capture new data-dependent invariances, the model architecture itself still suffers from a weak inductive bias.


In contrast, existing network pruning works found shallow MLPs can automatically be compressed into sparse subnetworks with good inductive bias by pruning the MLP itself
~\citep{neyshabur2020towards}. Combining pruning and invariance learning has largely been unsuccessful~\citep{corti2022studying}. Furthermore, pruning for invariance does not scale to deep MLPs, possibly due to issues in the lazy training regime~\citep{tzen2020mean,chizat2019lazy} where performance improves while weights magnitudes stay near static over training. Combining invariance learning with network pruning remains an open question.




\begin{figure*}
  \centering
  \includegraphics[width=0.7\textwidth]{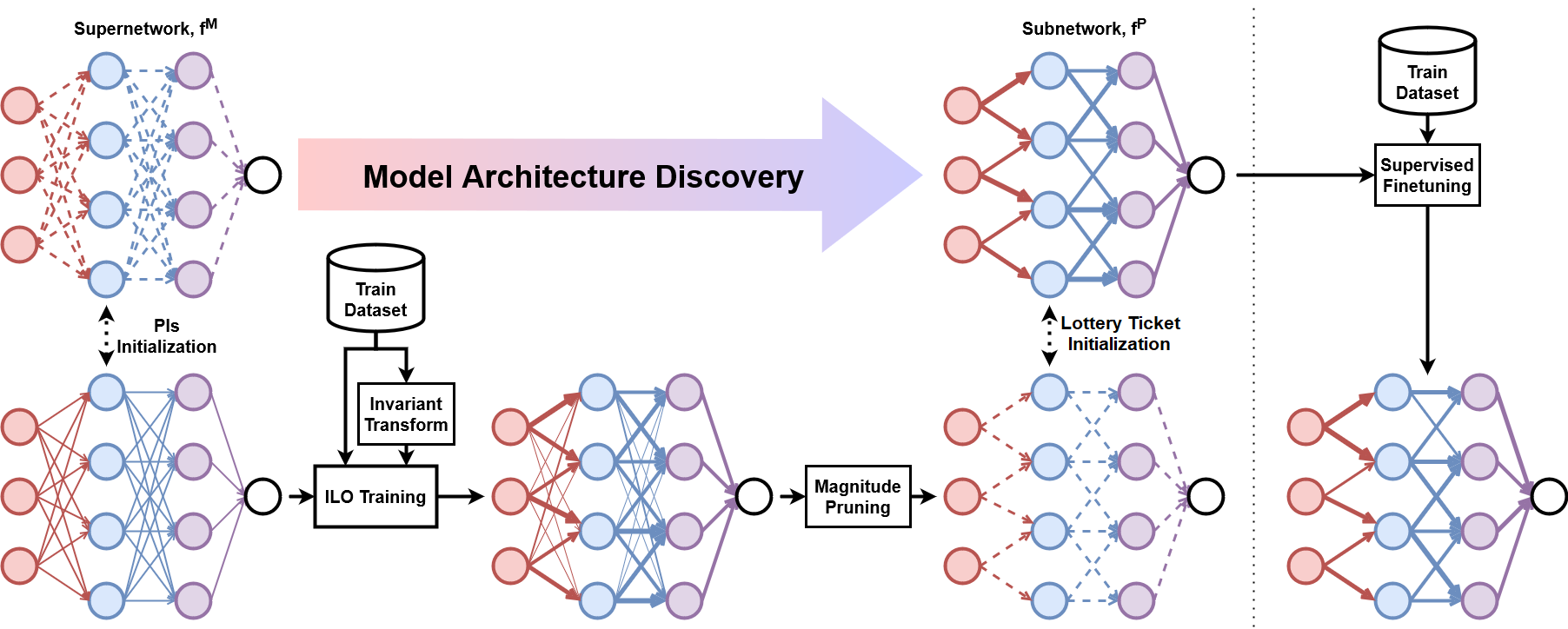}
  \caption{Overview for the \our Framework. The supernetwork, $f^P(\cdot,\theta_M)$, is initialized using \pis and trained on the \ilo objective to obtain $\theta_M^{(T)}$. Magnitude-based pruning is used to get a new architecture $f^P=\mathcal{P}(\theta_M^{(T)})$. The new architecture, $f^P(\cdot,\theta_P)$, is initialized via lottery ticket reinitialization and finetuned with supervised maximum likelihood loss.}
  \label{fig:framework}
\end{figure*}

We propose \underline{\textbf{I}}nvariance \underline{\textbf{U}}nveiling Neural \underline{\textbf{Net}}works, \our, a pruning framework that discovers invariance-preserving subnetworks from deep and dense supernetworks. We hypothesize pruning for invariances fails on deep networks due to the lazy training issue~\citep{liu2023sparsity}, where performance improvement decouples from weight magnitudes. We address this with a proactive initialization scheme (\pis), which prevents important weights from being accidentally pruned by assigning low magnitudes to majority of weights. To capture useful invariances, we propose a novel invariance learning objective (\ilo), that successfully combines CL with network pruning by regularizing CL with the supervised objective.

To the best of our knowledge, we are the first to automatically design deep architectures that incorporate invariance using pruning. We summarize our contributions below:

\begin{itemize}
  \item Designing architectures from scratch is difficult when desired invariances are either unknown or hard to incorporate. We automatically discover an invariance-preserving subnetwork that outperforms an invariance-agnostic supernetwork on both computer vision and tabular data. 
  \item  Network pruning is used to compress models for mobile devices. Our approach consistently improves compression performance for existing vision and tabular models.
  
  \item Contrastive learning traditionally fails with network pruning. We are the first to successfully combine contrastive learning with network pruning by regularizing it in our invariance learning objective.
  \item In the lazy training regime, performance improves drastically while weight magnitudes stay relatively constant, hence weights important for downstream performance may not have large magnitudes and hence be falsely pruned. We provide a simple yet effective approach that encourages only important weights to have large magnitudes before the lazy training regime begins.
\end{itemize}


%% file: sec-related.tex
\section{Related Work}

\subsection{Learning Invariances}

Most invariant networks are handcrafted for spatial invariances~\citep{dehmamy2021automatic,satorras2021n, deng2021vector, qi2017pointnet++, vaswani2017attention, cohen2016group, kipf2016semi, jaderberg2015spatial, lecun1998gradient}. Learning approaches primarily involve data augmentation followed by ensembling.~\citep{immer2022invariance,quiroga2020revisiting,lorraine2020optimizing,benton2020learning,cubuk2018autoaugment}. Some works use meta-learning to do parameter sharing in a given architecture~\citep{zhou2020meta,kirsch2022introducing}. None of the aforementioned works generates architectures from scratch to improve the network's inductive bias. The closest work is \betalasso~\citep{neyshabur2020towards} which discovers shallow subnetworks with local connectivity through pruning for computer vision. Our work extends to deeper networks and explores the tabular data setting. 



\begin{table*}
\centering
\begin{tabular}[width=0.5\textwidth]{|c | c | c c c |} 
 \hline
Dataset & \mlpv & \ompmlpv & \betalassomlpv & \ourmlpv \\
\hline
CIFAR10 & 59.266 $\pm$ 0.050 & 59.668 $\pm$ 0.171 & 59.349 $\pm$ 0.174 & {\bf 64.847 $\pm$ 0.121} \\
CIFAR100 & 31.052 $\pm$ 0.371 & 31.962 $\pm$ 0.113 & 31.234 $\pm$ 0.354 & {\bf 32.760 $\pm$ 0.288} \\
SVHN & 84.463 $\pm$ 0.393 & 85.626 $\pm$ 0.026 & 84.597 $\pm$ 0.399 & {\bf 89.357 $\pm$ 0.156} \\
\hline
\multicolumn{1}{c}{\vspace{0.5pt}}\\
 \hline
Dataset & \resnet & \ompres & \betalassores & \ourres \\
\hline
CIFAR10 & 73.939 $\pm$ 0.152 & 75.419 $\pm$ 0.290 & 74.166 $\pm$ 0.033 & {\bf 83.729 $\pm$ 0.153} \\
CIFAR100 & 42.794 $\pm$ 0.133 & 44.014 $\pm$ 0.163 & 42.830 $\pm$ 0.412 & {\bf 53.099 $\pm$ 0.243} \\
SVHN & 90.235 $\pm$ 0.127 & 90.474 $\pm$ 0.192 & 90.025 $\pm$ 0.201 & {\bf 94.020 $\pm$ 0.291} \\
\hline
\end{tabular}
\caption{Comparing different pruning approaches to improve the inductive bias of \mlpv and \resnet on computer vision datasets. Notice, \our performs substantially better than existing pruning-based methods by discovering novel architectures that better capture the inductive bias. \our flexibly boosts performance of off-the-shelf models.}
\label{tab:mlp_cv1} 
\end{table*}

\begin{table}
\centering
\begin{tabular}[width=0.5\textwidth]{|c | c c |} 
 \hline
Dataset & \ourmlpv & CNN \\
\hline
CIFAR10 & 64.847 $\pm$ 0.121 & {\bf 75.850 $\pm$ 0.788} \\
CIFAR100 & 32.760 $\pm$ 0.288 & {\bf 41.634 $\pm$ 0.402} \\
SVHN & 89.357 $\pm$ 0.156 & {\bf 91.892 $\pm$ 0.411} \\
\hline
\end{tabular}
\caption{Comparing the pruned \ourmlpv model to an equivalent CNN. Although \ourmlpv cannot outperform CNN, it bridges the gap between MLP and CNN architectures without any human design intervention.}
\label{tab:mlp_cv2} 
\end{table}

\subsection{Neural Network Pruning}

Neural network pruning compresses large supernetworks without hurting performance~\citep{frankle2018lottery, louizos2017learning}. A pinnacle work is the Lottery Ticket Hypothesis (LTH)~\citep{frankle2018lottery, liu2018rethinking, blalock2020state}, where pruned networks can retain unpruned peformance when reinitialized to the start of training and iteratively retrained. One-Shot Magnitude Pruning (OMP) studies how to prune the network only once~\citep{blalock2020state}. Recent work~\citep{liu2023sparsity} identifies the lazy training regime~\citep{chizat2019lazy} as a bottleneck for network pruning. Recent work~\citep{corti2022studying} find contrastive learning does not work with netwprl pruning. Recent pruning policies improve efficiency by starting with a sparse network~\citep{evci2020rigging}. or performing data-agnostic Zero-Shot Pruning~\citep{hoang2023revisiting, wang2020picking, lee2019signal}. Interestingly, subnetworks rarely outperform the original supernetwork, which has been dubbed the ``Jackpot'' problem~\citep{ma2021sanity}. In contrast to existing works, we successfully combine OMP with contrastive learning, alleviate the lazy learning issue, and outperform the original supernetwork.



%% file: sec-model.tex
\section{Proposed Method: \our}

\subsection{Problem Setting}

We study the classification task with inputs, $x \in \mathcal{X}$, class labels, $y \in \mathcal{Y}$, and hidden representations, $h\in\mathcal{H}$. Our neural network architecture, $f(x, \theta): \mathcal{X} \rightarrow \mathcal{Y}$ is composed of an encoder, $f_\mathcal{E}(\cdot, \theta): \mathcal{X} \rightarrow \mathcal{H}$ and decoder, $f_\mathcal{D}(\cdot, \theta): \mathcal{H} \rightarrow \mathcal{Y}$, where $\theta \in \Theta$ are the weights and  $f = f_\mathcal{E} \circ f_\mathcal{D}$. In the context of training, we denote the weights after $0<t<T$ iterations of stochastic gradient descent as $\theta^{(t)}$.

First, we define our notion of invariance. Given a set of invariant transformations, $\mathcal{S}$, we wish to discover a neural network architecture $f^*(x, \theta)$, such that all invariant input transformations map to the same representation, shown in Equation~\ref{eq:invar}. We highlight our task focuses on the discovery of novel architectures, $f^*(\cdot, \theta)$, not weights, $\theta$, because good architectures improves the inductive bias~\citep{neyshabur2017implicit}. 
\begin{equation}
\label{eq:invar}
f^*_\mathcal{E}(x, \theta) = f^*_\mathcal{E}(g(x), \theta) , \forall g \in \mathcal{S}, \forall \theta \in \Theta.
\end{equation}


\begin{table}
\centering
\begin{tabular}[width=0.5\textwidth]{| c | c | c c |} 
 \hline
Dataset & $g(\cdot)$ & \mlpv & \ourmlpv \\
\hline
\multirow{4}{*}{CIFAR10}
 & resize. & 44.096 $\pm$ 0.434 & {\bf 97.349 $\pm$ 4.590} \\
 & horiz. & 80.485 $\pm$ 0.504 & {\bf 99.413 $\pm$ 1.016} \\
 & color. & 56.075 $\pm$ 0.433 & {\bf 98.233 $\pm$ 3.060} \\
 & graysc. & 81.932 $\pm$ 0.233 & {\bf 99.077 $\pm$ 1.598} \\
\hline
\multicolumn{1}{c}{\vspace{0.5pt}}\\

 \hline
Dataset & $g(\cdot)$ & \mlpt & \ourmlpt \\
\hline
mfeat. & feat. & 46.093 $\pm$ 1.353 & {\bf 51.649 $\pm$ 4.282} \\
\hline
\end{tabular}
\caption{Comparing the consistency metric (\%) of the untrained supernetwork, \mlpv and \mlpt, against \our's pruned subnetwork under different invariant transforms, $g(\cdot)$. \our preserves invariances better.}
\label{tab:con} 
\end{table}

\subsection{Framework}

We accomplish this by first training a dense supernetwork, $f^{M}(\cdot,\theta_{M})$, with enough representational capacity to capture the desired invariance properties, as shown in Equation~\ref{eqn:super}. A natural choice for $f^{M}(\cdot,\theta_{M})$ is a deep MLP, which is a universal approximator~\citep{cybenko1989approximation}.

\begin{equation}
\label{eqn:super}
\exists \theta_M^* \in \Theta_M : f^{M}_\mathcal{E}(x, \theta_{M}^*) = f^{M}_\mathcal{E}(g(x), \theta_{M}^*) , \forall g \in \mathcal{S}.
\end{equation}

Next, we initialize the supernetwork's weights, $\theta_{M}^{(0)}$, using our Proactive Initialization Scheme, \pis, and train the supernetwork with our Invariance Learning Objective, \ilo, to obtain $\theta_{M}^{(T)}$. We discuss both \pis's and \ilo's details in following sections.

We construct our new untrained subnetwork, $f^P(\cdot, \theta_P^{(0)})$, from the trained supernetwork, $f^M(\cdot, \theta_M^{(T)})$, where the subnetwork contains a fraction of the supernetwork's weights, $\theta_P^{(0)} \subset \theta_M^{(T)}$ and $|\theta_P^{(0)}| \ll |\theta_M^{(T)}|$, and is architecturally different from the supernetwork, $f^P(\cdot, \cdot) \neq f^M(\cdot, \cdot)$. For this step, we adopt standard One-shot Magnitude-based Pruning (OMP), where the smallest magnitude weights and their connections in the supernetwork architecture are dropped. We adopt OMP because of its success in neural network pruning~\citep{frankle2018lottery,blalock2020state}. We represent this step as an operator mapping supernetwork weights into subnetwork architectures $\mathcal{P}: \Theta_M \rightarrow \mathcal{F}_P$, where $\mathcal{F}_P$ denotes the space of subnetwork architectures. 

We hypothesize the trained subnetwork, $f^P(\cdot, \theta_P^{(T)})$, can outperform the trained original supernetwork, $f^M(\cdot, \theta_M^{(T)})$, if it learns to capture the right invariances and hence achieving a better inductive bias. The ideal subnetwork, $f^{P*}(\cdot, \theta_{P*})$, could capture invariances without training, as shown in Equation~\ref{eqn:prune}.



\begin{equation}
\label{eqn:prune}
f^{P*}_\mathcal{E}(x, \theta_{P*}) = f^{P*}_\mathcal{E}(g(x), \theta_{P*}) , \forall g \in \mathcal{S}, \forall \theta_{P*} \in \Theta_{P*}
\end{equation}

Finally, we re-initialize the subnetwork's weights, $\theta_P^{(0)}$, using the Lottery Ticket Re-initialization scheme~\citep{frankle2018lottery} then finetune the subnetwork with maximum likelihood loss to obtain $\theta_P^{(T)}$. We call this framework, including the \ilo objective and \pis initialization scheme, \our\footnote{\our prunes an ineffective supernetwork into an efficient effective subnetwork. OMP prunes an inefficient effective supernetwork into an efficient but slightly less effective subnetwork.}, as shown in Figure~\ref{fig:framework}


\subsubsection{Invariance Learning Objective: \ilo}
\label{subsubsec:ilo}

The goal of supernetwork training is to create a subnetwork, $f^P(\cdot, \theta_P^{(0)})$, within the supernetwork, $f^M(\cdot, \theta_M^{(T)})$, such that:

\begin{enumerate}
  \item  $\mathcal{P}(\theta_M^{(T)})$ achieves superior performance on the classification task after finetuning.
  \item $\mathcal{P}(\theta_M^{(T)})$ captures desirable invariance properties as given by Equation~\ref{eqn:prune}.
  \item $\theta_P^{(0)}$ has higher weight values than $\theta_M^{(T)} \setminus \theta_P^{(0)}$.  
\end{enumerate}
 
Because subnetworks pruned from randomly initialized weights, $\mathcal{P}(\theta_M^{(0)})$, are poor, they include harmful inductive biases that hinders training. Thus, we optimize the trained supernetwork, $f^M(\cdot, \theta^{(T)}_M)$, on goals (1) and (2) as a surrogate training objective. Goal (3) is handled by \pis, described in the next section.

To achieve (1), we maximize the log likelihood of the data. To achieve (2), we minimize a distance metric, $\phi(\cdot, \cdot)$, between representations of inputs under invariant perturbations and maximize the metric between different input samples, given by Equation~\ref{eqn:obj}. We prove this is equivalent to Supervised Contrastive Learning (SCL) in the Supplementary Material. Hence, (2) can be achieved through SCL. 

\begin{equation}
\label{eqn:obj}
\theta_M^* = \mathop{argmax}_{\theta_M} \mathop{\mathbb{E}}_{\substack{x_i\\ x_j \sim \mathcal{X} \\ g \sim \mathcal{S}}} \left [ \frac{\phi ( f^{M}_\mathcal{E}(x_i, \theta_{M}), f^{M}_\mathcal{E}(x_j, \theta_{M}) )}{\phi ( f^{M}_\mathcal{E}(x_i, \theta_{M}), f^{M}_\mathcal{E}(g(x_i), \theta_{M}) )} \right ]
\end{equation}

Our final Invariance Learning Objective (\ilo) loss function combines these two ideas as shown in Equation~\ref{eqn:ilo}, where $\mathcal{L}_{SUP}$ is standard maximum likelihood loss, $\mathcal{L}_{NCE}$ is a contrastive loss (described in Appendix), $D_{tr}$ is a labelled training dataset of $(x,y)$ pairs, and $\lambda$ is a hyperparameter.

\begin{equation}
\label{eqn:ilo}
    \begin{gathered}
\mathcal{L}(\theta_M; \mathcal{S}) = \\ \mathop{\mathbb{E}}_{x,y \sim D_{tr}} \left [ \mathcal{L}_{SUP}(x, y, \theta_M) + \lambda \mathcal{L}_{NCE}(x, y, \theta_M; \mathcal{S}) \right ]
    \end{gathered}
\end{equation}

Both loss components are crucial to \our. With just $L_{NCE}$, the supernetwork will overfit the contrastive objective~\citep{corti2022studying, pasad2021layer}, causing weights critical for finetuning the supervised objective to be pruned. With just  $L_{SUP}$, the architecture is not explicitly optimized to capture desired invariances.


\subsubsection{Proactive Initialization Scheme: \pis}
\label{subsubsec:pis}

Deep neural networks often enter the lazy training regime~\citep{chizat2019lazy, liu2023sparsity}, where the loss steadily decreases while weights barely change. This is particularly harmful to neural networks pruning~\citep{liu2023sparsity}, especially when low-magnitude weights contribute to decreasing the loss and hence should not be pruned. 

We propose a simple solution by scaling the weight initialization by a small multiplier, $\kappa$. We find this alleviates the aforementioned issue by forcing the model to assign large values only to important weights prior to lazy training. Because lazy training is only an issue for pruning, we only apply $\kappa$-scaling to the pre-pruning training stage, not the fine-tuning stage. This is done by scaling the initial weights $\theta^{(0)}_M = \kappa \theta^{(0)}_{M^{\dagger}}$, where $\theta^{(0)}_{M^{\dagger}}$ follows the Kaiming~\citep{he2015delving} or Glorot~\citep{glorot2010understanding} initialization.


%% file: sec-exp.tex
\begin{table*}
\begin{center}
\begin{tabular}[width=0.5\textwidth]{|c | c | c c c c |} 
 \hline
Dataset & \mlpv & \ourmlpvP & \ourmlpvILO & \ourmlpvPIS & \ourmlpv \\
\hline
CIFAR10 & 59.266 $\pm$ 0.050 & 54.622 $\pm$ 0.378 & 62.662 $\pm$ 0.169 & 60.875 $\pm$ 0.292 & {\bf 83.729 $\pm$ 0.153} \\
CIFAR100 & 31.052 $\pm$ 0.371 & 20.332 $\pm$ 0.065 & 32.242 $\pm$ 0.321 & 32.747 $\pm$ 0.346 & {\bf 53.099 $\pm$ 0.243} \\
SVHN & 84.463 $\pm$ 0.393 & 78.427 $\pm$ 0.683 & 88.870 $\pm$ 0.139 & 85.247 $\pm$ 0.071 & {\bf 94.020 $\pm$ 0.291} \\
\hline
\multicolumn{1}{c}{\vspace{0.5pt}}\\
 \hline
Dataset & \mlpt & \ourmlptP & \ourmlptILO & \ourmlptPIS & \ourmlpt \\
\hline
arrhythmia & 67.086 $\pm$ 2.755&56.780 $\pm$ 6.406&71.385 $\pm$ 6.427&{\bf 78.675 $\pm$ 7.078}&74.138 $\pm$ 2.769 \\
mfeat. & 98.169 $\pm$ 0.297&97.528 $\pm$ 0.400&{\bf 98.471 $\pm$ 0.344}&98.339 $\pm$ 0.203&98.176 $\pm$ 0.121 \\
vehicle & 80.427 $\pm$ 1.829&80.427 $\pm$ 1.806&81.411 $\pm$ 0.386&80.928 $\pm$ 0.861&{\bf 81.805 $\pm$ 2.065} \\
kc1 & 80.762 $\pm$ 1.292&{\bf 84.597 $\pm$ 0.000}&82.456 $\pm$ 1.850&{\bf 84.597 $\pm$ 0.000}&{\bf 84.597 $\pm$ 0.000} \\
\hline
\end{tabular}
\caption{Ablation Study on vision and tabular datasets.}
\label{tab:abl} 
\vspace{-12pt}
\end{center}
\end{table*}

\begin{figure}
\begin{tabular}{cc}
\subfloat[CIFAR10 \mlpv (\pis)]{\includegraphics[width = 1.5in]{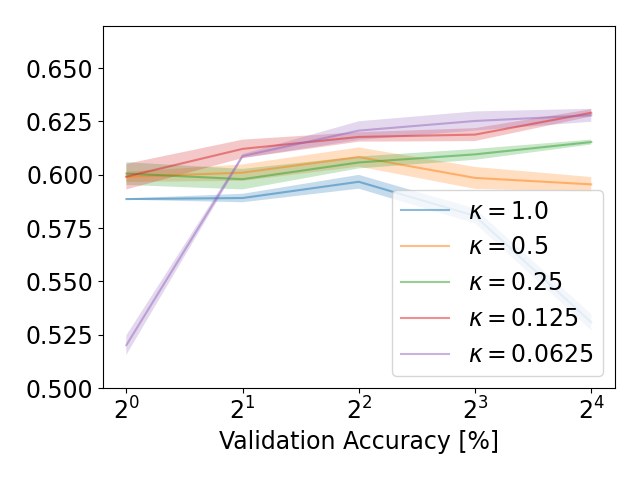}} &
\subfloat[CIFAR10 \mlpv (\ilo)]{\includegraphics[width = 1.5in]{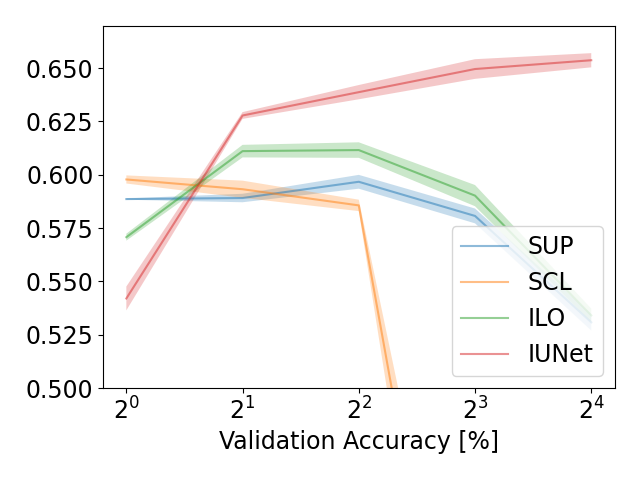}} \\
\subfloat[CIFAR100 \mlpv (\pis)]{\includegraphics[width = 1.5in]{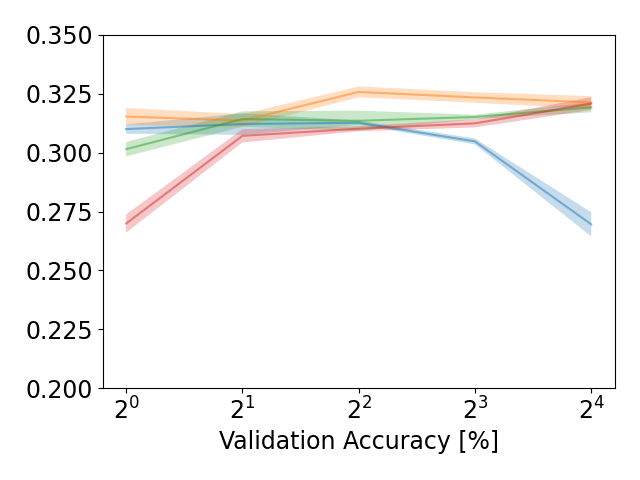}} &
\subfloat[CIFAR100 \mlpv (\ilo)]{\includegraphics[width = 1.5in]{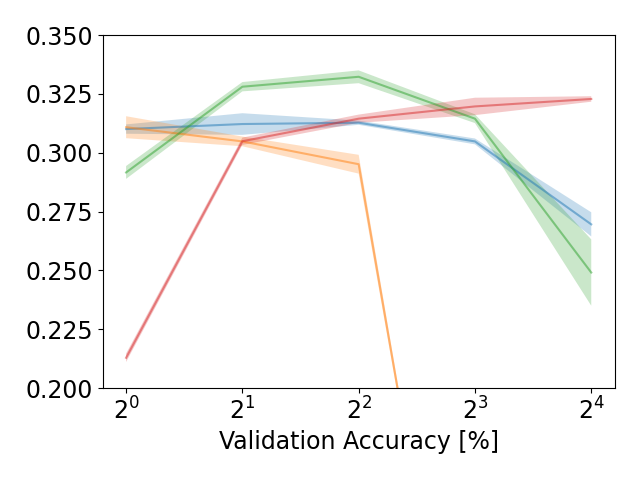}} \\ 
\subfloat[SVHN \mlpv (\pis)]{\includegraphics[width = 1.5in]{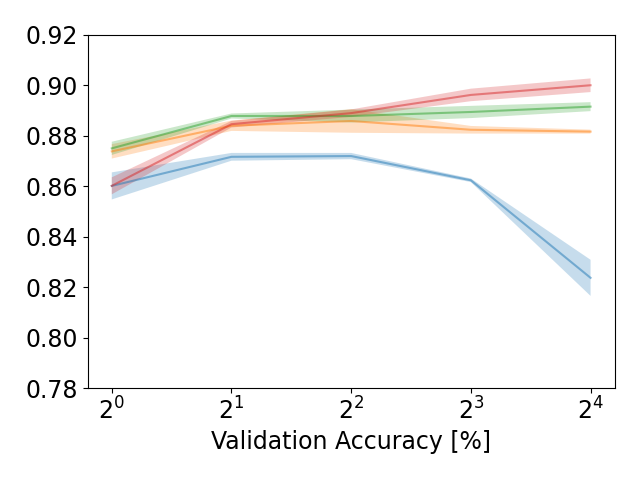}} &
\subfloat[SVHN \mlpv (\ilo)]{\includegraphics[width = 1.5in]{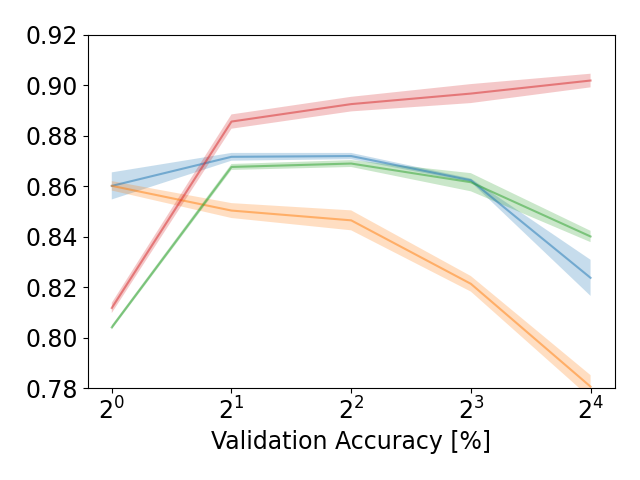}}
\end{tabular}
\caption{Effect of \pis and \ilo on pruned models. The y-axis is the validation accuracy (\%) and x-axis is the compression ratio. \pis experiments only alter the supernetwork's initialization. $\kappa=1.0$ means normal initialization. \ilo experiments only alter the training objective during supernetwork training. After supernetwork training, subnetworks are pruned under different compression ratios, then finetuned. Validation accuracy of trained pruned models are reported.}
\label{fig:prune}
\end{figure}

\section{Experiment Setup}

\subsection{Datasets}
\our is evaluated on {\it image} and {\it tabular} classification \footnote{More details are provided in the Supplementary.}:

\begin{itemize}
  \item {\bf Vision}: Experiments are run on CIFAR10, CIFAR100, and SVHN~\citep{krizhevsky2009learning, netzer2011reading}, following baseline work~\citep{neyshabur2020towards}\footnote{While SMC benchmark~\citep{liu2023sparsity} is open-sourced, it is undergoing code clean-up at the time of submission.}.
 
  \item {\bf Tabular}: Experiments are run on 40 tabular datasets from a benchmark paper~\citep{kadra2021well}, covering a diverse range of problems. The datasets were collected from OpenML~\citep{gijsbers2019open}, UCI~\citep{asuncion2007uci}, and Kaggle.
\end{itemize}

\subsection{Model Setup}
\our is compared against One-shot Magnitude Pruning (OMP)~\citep{blalock2020state}, and \betalasso pruning~\citep{neyshabur2020towards} on all datasets. We denote the supernetwork used by each pruning method with a superscript. Unless otherwise specified, models are trained via maximum likelihood. In addition, we compare against the following dataset-specific supernetworks (\mlpv, \mlpt, \resnet) and models:

\begin{itemize}
  \item {\bf Vision}: We consider \resnet~\citep{he2016deep}, \mlpv, a MLP that contains a CNN subnetwork~\citep{neyshabur2020towards}, and the aforementioned CNN subnetwork.

  \item {\bf Tabular}: We consider \mlpt, a 9-layer MLP with hidden dimension 512~\citep{kadra2021well}, \xgb~\citep{chen2016xgboost}, \tabn~\citep{arik2021tabnet}, a handcrafted tabular deep learning architecture, and \mlpc~\citep{kadra2021well}, the state-of-the-art MLP, which was heavily tuned from a cocktail of regularization techniques.
\end{itemize}

\subsection{Considered Invariances}
Because contrastive learning is successful on both vision and tabular datasets, our invariant transformations, $\mathcal{S}$, come from existing works. For computer vision, SimCLR~\citep{chen2020simple} transformations are used: (1) resize crops, (2) horizontal flips, (3) color jitter, and (4) random grayscale. For tabular learning, SCARF~\citep{bahri2021scarf} transformations are used: (5) randomly corrupting features by drawing the corrupted versions from its empirical marginal distribution.




%% file: sec-result.tex
\section{Results}

\subsection{On Inductive Bias}

In this section, we compare the effectiveness of the trained subnetwork discovered by \our, $f^P(\cdot, \theta^{(T)}_P)$, against the trained supernetwork, $f^M(\cdot, \theta^{(T)}_M)$. As seen in Tables~\ref{tab:mlp_cv1} and~\ref{tab:tab}, the pruned subnetwork outperforms the original supernetwork, even though the supernetwork has more representational capacity. This supports our claim that \our prunes subnetwork architectures with better inductive biases than the supernetwork. Importantly, \our substantially improves upon existing pruning baselines by explicitly including invariances via \ilo and alleviating the lazy learning issue~\citep{liu2023sparsity} via \pis.

On  {\it vision} datasets: As seen in Figure~\ref{tab:mlp_cv1}, \our is a general and flexible framework that improves the inductive bias of not only models like \mlpv but also specialized architectures like \resnet. As seen in Figure~\ref{tab:mlp_cv2}, \our bridges the gap between MLPs and CNNs. Unlike previous work~\citep{tolstikhin2021mlp}, \our does this in an entirely automated procedure. While CNN outperforms \ourmlpv, we can also apply \our to specialized networks, \ourres, which achieves the best overall performance. Figures~\ref{tab:mlp_cv1} and~\ref{tab:mlp_cv2} show \our is a useful tool for injecting inductive bias into arbitrary neural architectures.

On {\it tabular} datasets: As seen in Figure~\ref{tab:tab}, the subnetworks derived from MLPs outperform both the original \mlpt and hand-crafted architectures: \tabn and \xgb. Unlike vision, how to encode invariances for tabular data is highly nontrivial, making \our particularly effective. Note, although \our performs competitively against \mlpc, they are orthogonal approaches. \mlpc focuses on tuning regularization hyperparameters during finetuning, whereas \our improves the model architecture. Note, \ourmlpt did not use the optimal hyperparameters found by \mlpc~\citep{kadra2021well}.

\subsection{Ablation Study}
To study the effectiveness of (1) pruning, (2) \pis, and (3) \ilo, each one is removed from the optimal model. As seen in Table~\ref{tab:abl}, each is crucial to \our. Pruning is necessary to encode the inductive bias into the subnetwork's neural architecture. \pis and \ilo improves the pruning policy by ensuring weights crucial to finetuning and capturing invariance are not pruned. Notice, without pruning, \ourP performs worse than the original supernetwork. This highlights an important notion that \pis aims to improve the pruning policy, not the unpruned performance. By sacrificing unpruned performance, \pis ensures important weights are not falsely pruned. \pis is less effective on tabular datasets where the false pruning issue seems less severe. Combining pruning, \ilo, and \pis, \our most consistently achieves the best performance.

\subsection{Effects of Pruning}
To further study the effects of pruning, we plot how performance changes over different compression ratios. Figure~\ref{fig:prune} clearly identifies how \pis and \ilo substantially improves upon existing pruning policies. First, our results support existing findings that (1) OMP does not produce subnetworks that substantially outperform the supernetwork~\citep{blalock2020state} and (2) while unpruned models trained with SCL can outperform supervised ones, pruned models trained with SCL perform substantially worse~\citep{corti2022studying}. \pis flips the trend from (1) by slightly sacrificing unpruned performance, due to poorer initialization, \our discovers pruned models with better inductive biases, which improves downstream performance. \ilo fixes the poor performance of SCL in (2) by preserving information pathways for both invariance and max likelihood over training. We highlight both these findings are significant among the network pruning community. Finally, Figure~\ref{fig:prune} confirms \our achieves the best performance by combining both \pis and \ilo.

In addition to being more effective that the supernetwork, $f^M(\cdot, \theta^{(T)}_M)$, the pruned network, $f^P(\cdot, \theta^{(T)}_P)$, is also more efficient. Figure~\ref{fig:prune} shows \our can reach 8-16$\times$ compression while still keeping superior performance.

\begin{figure}
\begin{tabular}{cc}
\subfloat[$\kappa=1.0$]{\includegraphics[width = 1.5in]{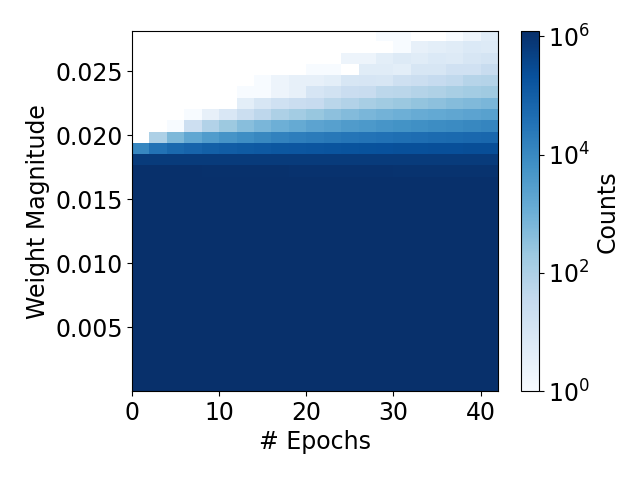}} &
\subfloat[$\kappa=0.125$]{\includegraphics[width = 1.5in]{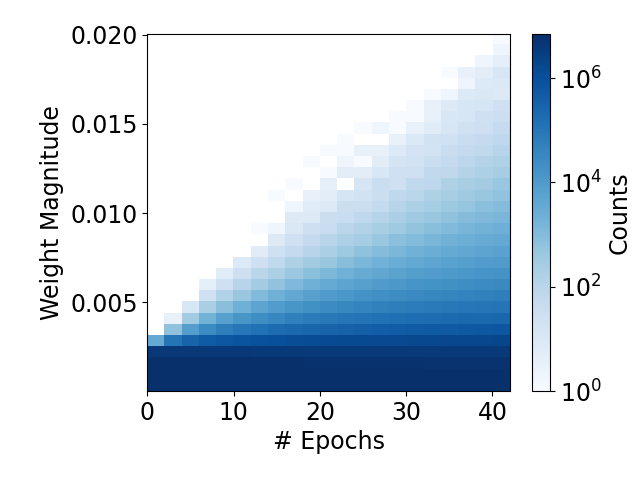}}
\end{tabular}
\caption{Histogram of weight magnitudes, $|\theta_M^{(t)}|$, plotted over each epoch under different $\kappa$ initializations settings. $\kappa=1.0$ means normal initialization. Results shown for \mlpv on the CIFAR10 dataset.}
\label{fig:pis}
\end{figure}

\begin{figure}
\begin{tabular}{cc}
\subfloat[\mlpv]{\includegraphics[width = 1.5in]{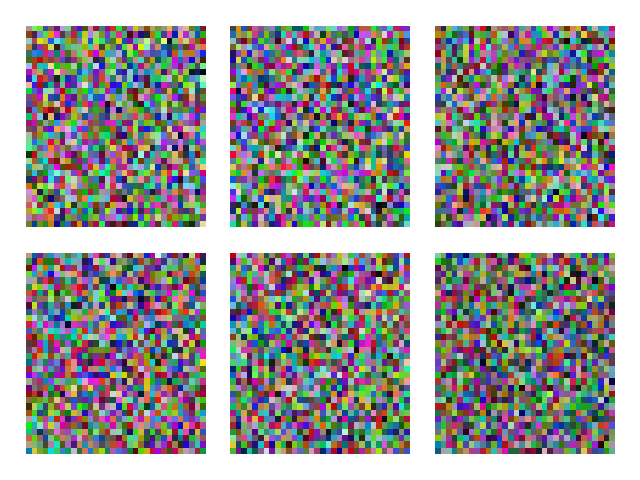}} &
\subfloat[\ourmlpv]{\includegraphics[width = 1.5in]{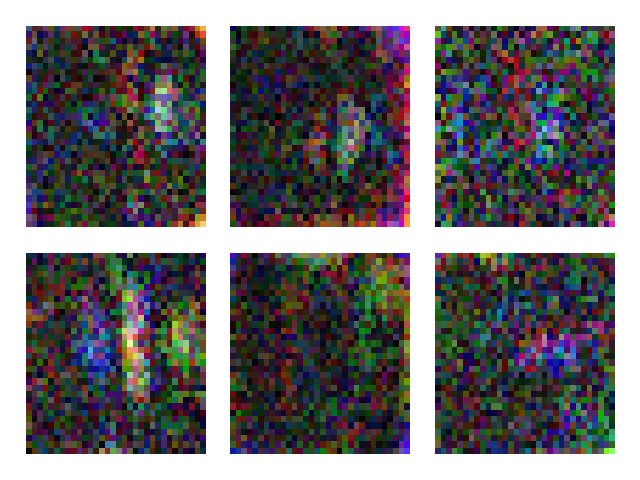}} \\
\subfloat[\mlpt]{\includegraphics[width = 1.4in]{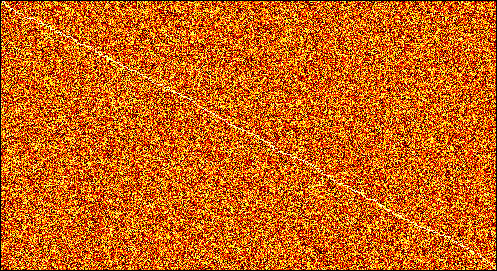}} &
\subfloat[\ourmlpt]{\includegraphics[width = 1.4in]{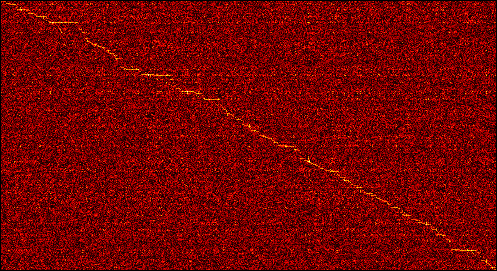}}
\end{tabular}
\caption{Visualization of weight magnitudes, $|\theta_M^{(T)}|$, trained with different policies. The top row was trained on CIFAR10 and shows the magnitude of each RGB pixel for 6 output logits. The bottom row was trained on arrhythmia and shows the weight matrix of the 1st layer with 280 input and 512 output dimensions. Lighter color means larger magnitude.}
\label{fig:vis}
\end{figure}

\subsection{Effect of Proactive Initialization}
To further study the role of \pis, the histogram of weight magnitudes is monitored over the course of training. As shown in Figure~\ref{fig:pis}, under the standard OMP pruning setup, the histogram changes little over the course of training, which supports the lazy training hypothesis~\citep{liu2023sparsity} where performance rapidly improves, while weight magnitudes change very little, decoupling each weight's importance from its magnitude. 

With \pis, only important weights grow over the course of training, while most weights remain near zero, barely affecting the output activations of each layer. This phenomenon alleviates the lazy training problem by ensuring (1) pruning safety, as pruned weights are near zero prior which have minimal affect on layer activations, and (2) importance-magnitude coupling, as structurally important connections must grow to affect the output of the layer.

\subsection{On Invariance Consistency}
To further study whether particular invariances are learned, we compute the consistency metric~\citep{singla2021shift}, which measure the percentage of samples whose predicted label would flip when an invariant transformation is applied to the input. As seen in Table~\ref{tab:con}, the subnetwork found by \our, $f^P(\cdot, \theta^{(0)}_P)$, is able to preserve invariances specified in \ilo much better than the supernetwork, $f^M(\cdot, \theta^{(0)}_M)$. This shows \our indeed captures desirable invariances.

\subsection{On Weight Visualization}
We visualize the supernetwork weights, $\theta^{(T)}_M$, when trained with \our compared to standard maximum likelihood (MLP) to determine what structures preserve invariance.

On  {\it vision} datasets: As seen in Figure~\ref{fig:vis}, \our learns more locally connected structures, which improves translation invariance. Prior work~\citep{neyshabur2020towards} found network structure (as opposed to inductive bias) to be the limiting factor for encoding CNN inductive biases into MLPs, which \our successfully replicates.

On {\it tabular} datasets: As seen in Figure~\ref{fig:vis}, \our weights focus more on singular features. This preserves invariance over random feature corruption, as the absence of some tabular features does not greatly alter output activations of most neurons. This structure can also be likened to tree ensembles~\citep{grinsztajn2022tree}, whose leaves split individual features rather than all features.


\begin{table*}
\begin{center}
\begin{tabular}[width=0.5\textwidth]{|c | c | c c c | c c c |} 
 \hline
Dataset & \mlpt & \ompmlpt & \betalassomlpt & \ourmlpt & \xgb & \tabn & \mlpc \\
\hline 
credit-g & 70.000&70.000 $\pm$ 0.000&67.205 $\pm$ 0.718&63.166 $\pm$ 0.000&68.929&61.190&{\bf74.643} \\
anneal & 99.490&99.691 $\pm$ 0.234&99.634 $\pm$ 0.518&{\bf 99.712 $\pm$ 0.101}&85.416&84.248&89.270 \\
kr-vs-kp & 99.158&99.062 $\pm$ 0.142&99.049 $\pm$ 0.097&99.151 $\pm$ 0.064&{\bf99.850}&93.250&{\bf99.850} \\
arrhythmia & 67.086&55.483 $\pm$ 5.701&67.719 $\pm$ 3.483&{\bf 74.138 $\pm$ 2.769}&48.779&43.562&61.461 \\
mfeat. & 98.169&97.959 $\pm$ 0.278&97.204 $\pm$ 0.620&{\bf 98.176 $\pm$ 0.121}&98.000&97.250&98.000 \\
vehicle & 80.427&81.115 $\pm$ 2.214&80.611 $\pm$ 1.244&81.805 $\pm$ 2.065&74.973&79.654&{\bf82.576} \\
kc1 & 80.762&{\bf 84.597 $\pm$ 0.000}&83.587 $\pm$ 1.010&{\bf 84.597 $\pm$ 0.000}&66.846&52.517&74.381 \\
adult & 81.968&82.212 $\pm$ 0.582&82.323 $\pm$ 0.424&78.249 $\pm$ 3.085&79.824&77.155&{\bf82.443} \\
walking. & 58.466&60.033 $\pm$ 0.112&58.049 $\pm$ 0.309&59.789 $\pm$ 0.456&61.616&56.801&{\bf63.923} \\
phoneme & 84.213&86.733 $\pm$ 0.194&84.850 $\pm$ 1.548&87.284 $\pm$ 0.436&{\bf87.972}&86.824&86.619 \\
skin-seg. & 99.869&99.866 $\pm$ 0.016&99.851 $\pm$ 0.015&99.876 $\pm$ 0.006&{\bf99.968}&99.961&99.953 \\
ldpa & 66.590&68.458 $\pm$ 0.140&62.362 $\pm$ 4.605&64.816 $\pm$ 4.535&{\bf99.008}&54.815&68.107 \\
nomao & 95.776&95.682 $\pm$ 0.046&95.756 $\pm$ 0.074&95.703 $\pm$ 0.110&{\bf96.872}&95.425&96.826 \\
cnae & 94.080&92.742 $\pm$ 0.404&94.808 $\pm$ 0.254&{\bf 96.075 $\pm$ 0.242}&94.907&89.352&95.833 \\
blood. & 68.965&61.841 $\pm$ 10.012&65.126 $\pm$ 20.792&{\bf 70.375 $\pm$ 5.255}&62.281&64.327&67.617 \\
bank. & {\bf88.300}&{\bf 88.300 $\pm$ 0.000}&86.923 $\pm$ 1.948&{\bf 88.300 $\pm$ 0.000}&72.658&70.639&85.993 \\
connect. & 72.111&72.016 $\pm$ 0.112&72.400 $\pm$ 0.214&74.475 $\pm$ 0.445&72.374&72.045&{\bf80.073} \\
shuttle & 99.709&93.791 $\pm$ 3.094&99.687 $\pm$ 0.027&93.735 $\pm$ 2.303&98.563&88.017&{\bf99.948} \\
higgs & 72.192&72.668 $\pm$ 0.039&72.263 $\pm$ 0.149&73.215 $\pm$ 0.384&72.944&72.036&{\bf73.546} \\
australian & 82.153&83.942 $\pm$ 1.578&81.667 $\pm$ 1.572&82.562 $\pm$ 1.927&{\bf89.717}&85.278&87.088 \\
car & 99.966&{\bf 100.000 $\pm$ 0.000}&{\bf 100.000 $\pm$ 0.000}&99.859 $\pm$ 0.200&92.376&98.701&99.587 \\
segment & 91.504&91.603 $\pm$ 0.508&91.317 $\pm$ 0.074&91.563 $\pm$ 0.000&{\bf93.723}&91.775&{\bf93.723} \\
fashion. & 91.139&90.784 $\pm$ 0.158&90.864 $\pm$ 0.090&90.817 $\pm$ 0.040&91.243&89.793&{\bf91.950} \\
jungle. & 86.998&92.071 $\pm$ 0.420&87.400 $\pm$ 0.489&95.130 $\pm$ 0.807&87.325&73.425&{\bf97.471} \\
numerai & 51.621&51.443 $\pm$ 0.370&51.905 $\pm$ 0.299&51.839 $\pm$ 0.067&52.363&51.599&{\bf52.668} \\
devnagari & 97.550&97.573 $\pm$ 0.031&97.549 $\pm$ 0.014&97.517 $\pm$ 0.014&93.310&94.179&{\bf98.370} \\
helena & 29.342&28.459 $\pm$ 0.531&29.834 $\pm$ 0.354&{\bf 29.884 $\pm$ 0.991}&21.994&19.032&27.701 \\
jannis & 68.647&66.302 $\pm$ 3.887&69.302 $\pm$ 0.248&{\bf 69.998 $\pm$ 1.232}&55.225&56.214&65.287 \\
volkert & 70.066&68.781 $\pm$ 0.045&69.655 $\pm$ 0.189&70.104 $\pm$ 0.215&64.170&59.409&{\bf71.667} \\
miniboone & 86.539&87.575 $\pm$ 0.855&87.751 $\pm$ 0.398&81.226 $\pm$ 6.569&{\bf94.024}&62.173&94.015 \\
apsfailure & 97.041&{\bf 98.191 $\pm$ 0.000}&98.048 $\pm$ 0.203&{\bf 98.191 $\pm$ 0.000}&88.825&51.444&92.535 \\
christine & 70.295&69.819 $\pm$ 0.462&70.275 $\pm$ 1.045&69.065 $\pm$ 1.225&{\bf74.815}&69.649&74.262 \\
dilbert & 98.494&98.738 $\pm$ 0.029&98.522 $\pm$ 0.084&98.540 $\pm$ 0.023&{\bf99.106}&97.608&99.049 \\
fabert & 65.540&64.709 $\pm$ 0.293&66.681 $\pm$ 0.208&65.695 $\pm$ 0.065&{\bf70.098}&62.277&69.183 \\
jasmine & 78.691&80.139 $\pm$ 1.978&78.415 $\pm$ 1.731&{\bf 80.864 $\pm$ 0.374}&80.546&76.690&79.217 \\
sylvine & 92.660&92.650 $\pm$ 0.267&92.593 $\pm$ 0.368&93.369 $\pm$ 0.833&{\bf95.509}&83.595&94.045 \\
dionis & 93.920&93.687 $\pm$ 0.059&93.943 $\pm$ 0.037&93.586 $\pm$ 0.021&91.222&83.960&{\bf94.010} \\
aloi & 96.546&96.376 $\pm$ 0.069&96.562 $\pm$ 0.087&95.341 $\pm$ 0.194&95.338&93.589&{\bf97.175} \\
ccfraud & 97.554&97.748 $\pm$ 1.622&96.626 $\pm$ 3.202&{\bf 98.797 $\pm$ 1.031}&90.303&85.705&92.531 \\
clickpred. & 82.175&83.206 $\pm$ 0.000&82.307 $\pm$ 0.500&{\bf 85.270 $\pm$ 1.275}&58.361&50.163&64.280 \\
\hline
\#Best & 1 & 4 & 1 & 13 & 12 & 0 & 16\\
\hline
\end{tabular}
\caption{Comparing \our against trees (\xgb), handcrafted models (\tabn), and state-of-the-art regularized MLPs (\mlpc). The last row denotes the number of datasets where the model achieves the best performance. Note, our method does not tune the optimal regularization settings for each dataset making it more efficient. Our pruned model is also more compressed than the original network. Note, we outperform both \mlpt and \tabn on most datasets. While \our performs similarly to \mlpc, it does not require costly hyperparameter tuning, and can be applied on top of the optimal settings found by \mlpc.}
\label{tab:tab} 
\end{center}
\end{table*}

%% file: sec-conclusion.tex
\section{Conclusion}

In this work, we study the viability of network pruning for discovering invariant-preserving architectures. Under the computer vision setting, \our bridges the gap between deep MLPs and deep CNNs, and reliably boosts \resnet performance. Under the tabular setting, \our reliably boosts performance of existing MLPs, comparable to applying the state-of-the-art regularization cocktails. Our proposed novelties, \ilo and \pis, flexibly improves existing OMP pruning policies by both successfully integrating contrastive learning and alleviating lazy training. Thus, \our effectively uses pruning to tackle invariance learning.

%% file: sec-supp-related.tex
\section{Additional Related Work}

\subsection{Tabular Machine Learning}

Tabular data is a difficult regime for deep learning, where deep learning models struggle against decision tree approaches. Early methods use forests, ensembling, and boosting~\citep{shwartz2022tabular, borisov2022deep, chen2016xgboost}. Later, researchers handcrafted new deep architectures that mimic trees~\citep{popov2019neural, arik2021tabnet}. Yet, when evaluated on large datasets, these approaches are still beaten by \xgb~\citep{chen2016xgboost, grinsztajn2022tree}. Recent work found MLPs with heavy regularization tuning~\citep{kadra2021well} can outperform decision tree approaches, though this conclusion does not hold on small tabular datasets~\citep{joseph2022gate}. To specially tackle the small data regime, Bayesian learning and Hopfield networks are combined with MLPs~\citep{hollmann2022tabpfn,schafl2022hopular}. There are also work on tabular transformers~\citep{huang2020tabtransformer}, though said approaches require much more training data. Without regularization, tree based models still outperform MLPs due to a better inductive bias and resilience to noise~\citep{grinsztajn2022tree}. To the best of our knowledge, the state-of-the-art on general tabular datasets remain heavily regularized MLPs (\mlpc)~\citep{kadra2021well}. We aim to further boost regularized MLP performance by discovering model architectures that capture good invariances from tabular data.

\subsection{Contrastive Learning}

Contrastive learning, initially proposed for metric learning~\citep{chopra2005learning,schroff2015facenet,oh2016deep}, trains a model to learn shared features among images of the same type~\citep{jaiswal2020survey}. It has been widely used in self-supervised pretraining~\citep{chen2020simple, chen2021empirical}, where dataset augmentation is crucial. Although contrastive learning was originally proposed for images, it has also shown promising results in  graph data~\citep{zhu2021graph,you2020graph}, speech data~\citep{baevski2020wav2vec}, and tabular data~\citep{bahri2021scarf}. Previous study has showed that speech transformers tend to overfit the contrastive loss in deeper layers, suggesting that removing later layers can be beneficial during finetuning~\citep{pasad2021layer}. While contrastive learning performs well pretraining unpruned models, its vanilla formulation performs poorly after network pruning~\citep{corti2022studying}. In this work, we establish a connection between contrastive learning and invariance learning and observe that pruned contrastive models fail because of overfitting.

\subsection{Neural Architecture Search}

Neural Architecture Search (NAS) explores large superarchitectures by leveraging smaller block architectures~\citep{wan2020fbnetv2,pham2018efficient, zoph2018learning, luo2018neural, liu2018darts}. These block architectures are typically small convolutional neural networks (CNNs) or MLPs. The key idea behind NAS is to utilize these blocks~\citep{pham2018efficient, zoph2018learning} to capture desired invariance properties for downstream tasks. Prior works~\citep{you2020graph2, xie2019exploring} have analyzed randomly selected intra- and inter-block structures and observed performance differences between said structures. However, these work did not propose a method for discovering block architectures directly from data. Our work aims to address this gap by focusing on discovering the architecture within NAS blocks.  This approach has the potential to enable NAS in diverse domains, expanding its applicability beyond the current scope.




%% file: sec-supp-loss.tex
\section{Loss Function Details}

We provide a more detailed description of our loss function in this section. Following notation from the main paper, we repeat the \ilo loss function in Equation~\ref{eqn:ilo} below:

\begin{equation}
\label{eqn:ilo}
\mathcal{L}(\theta; \mathcal{S}) = \mathbb{E}_{x,y \sim D_{train}} \left [ \mathcal{L}_{SUP}(x, y, \theta) + \lambda \mathcal{L}_{NCE}(x, y, \theta; \mathcal{S}) \right ]
\end{equation}

To better explain our loss functions, we introduce some new notations. First, we denote the decoder output probability function over classes, $\mathcal{Y}$, as $\tilde{p}_{\theta_\mathcal{D}}: \mathcal{H} \rightarrow [0,1]^{|\mathcal{Y}|}$, where $f_{\mathcal{D}} =  \text{argmax} \circ \tilde{p}_{\theta_\mathcal{D}}$. We denote the model output probability function by combining $\tilde{p}_{\theta_\mathcal{D}}$ with the encoder as follows: $p_{\theta} = \tilde{p}_{\theta_\mathcal{D}} \circ f_{\mathcal{E}}$. We introduce an integer mapping from classes $\mathcal{Y}$ as $\mathcal{I}: \mathcal{Y} \rightarrow \{0,1,2,...,|\mathcal{Y}|-1\}$.

We show the maximum likelihood loss, $\mathcal{L}_{SUP}$, in Equation~\ref{eqn:sup} below.

\begin{equation}
\label{eqn:sup}
\mathcal{L}_{SUP}(x,y,\theta) = -log(p_{\theta}(x, \theta)_{\mathcal{I}(y)})
\end{equation}

We show the supervised contrastive loss, $\mathcal{L}_{NCE}$, in Equation~\ref{eqn:nce} below. Following SimCLR~\citep{chen2020simple}, we assume that the intermediary representations are $d$-dimensional embeddings, $\mathcal{H} = \mathbb{R}^d$, and use the cosine similarity as our similarity function, $\psi^{(cos)}: \mathbb{R}^d \times \mathbb{R}^d \rightarrow \mathbb{R}$.

\begin{equation}
\label{eqn:nce}
\begin{split}
\mathcal{L}_{NCE}(x,y,\theta;\mathcal{S}) = \\
\mathop{\mathbb{E}}_{g \sim \mathcal{S}} \left [ -log \left ( \frac{exp \left ( \psi^{(cos)} \left ( \substack{f_{\mathcal{E}}(x, \theta), \\ f_{\mathcal{E}}(g(x), \theta)}\right ) \right )}{
\sum\limits_{\substack{x',y' \sim D_{tr} \\ y' \neq y}}^{}
exp \left ( \psi^{(cos)} \left ( \substack{f_{\mathcal{E}}(x, \theta), \\ f_{\mathcal{E}}(g(x'), \theta)} \right ) \right )
} \right ) \right]
\end{split}
\end{equation}

\subsection{Surrogate Objective}
We aim to learn invariance-preserving network architectures from the data. In our framework, this involves optimizing our invariance objective, which we repeat in Equation~\ref{eqn:obj}. We now prove that by minimizing the supervised contrastive loss in Equation~\ref{eqn:nce} we equivalently maximize the invariance objective, outlined below.

\begin{equation}
\label{eqn:obj}
\theta^* = \mathop{argmax}_{\theta} \mathop{\mathbb{E}}_{\substack{x_i, x_j \sim \mathcal{X} \\ g \sim \mathcal{S}}} \left [ \frac{\phi ( f^{M}_\mathcal{E}(x_i, \theta), f^{M}_\mathcal{E}(x_j, \theta) )}{\phi ( f^{M}_\mathcal{E}(x_i, \theta), f^{M}_\mathcal{E}(g(x_i), \theta) )} \right ]
\end{equation}

We convert the distance metric $\phi$ into similarity metric $\psi$.

\begin{equation}
\begin{split}
\theta^* &= \mathop{argmax}_{\theta} \mathop{\mathbb{E}}_{\substack{x_i, x_j \sim \mathcal{X} \\ g \sim \mathcal{S}}} \left [ \frac{\psi ( f^{M}_\mathcal{E}(x_i, \theta), f^{M}_\mathcal{E}(g(x_i), \theta) )}{\psi ( f^{M}_\mathcal{E}(x_i, \theta), f^{M}_\mathcal{E}(x_j, \theta) )} \right ]
\\
&= \mathop{argmin}_{\theta} \mathop{\mathbb{E}}_{\substack{x_i, x_j \sim \mathcal{X} \\ g \sim \mathcal{S}}} \left [ \frac{-\psi ( f^{M}_\mathcal{E}(x_i, \theta), f^{M}_\mathcal{E}(g(x_i), \theta) )}{\psi ( f^{M}_\mathcal{E}(x_i, \theta), f^{M}_\mathcal{E}(x_j, \theta) )} \right ]
\\
&= \mathop{argmin}_{\theta} \mathop{\mathbb{E}}_{\substack{x\sim \mathcal{X} \\ g \sim \mathcal{S}}}
\left [\frac{-\psi ( f^{M}_\mathcal{E}(x, \theta), f^{M}_\mathcal{E}(g(x), \theta) )}{\sum\limits_{\substack{x' \sim \mathcal{X} \\ x' \neq x}}
\psi(f_{\mathcal{E}}^M(x, \theta), f_{\mathcal{E}}^M(g(x'), \theta))
} \right ]
\\
&= \mathop{argmin}_{\theta} \mathop{\mathbb{E}}_{\substack{x, y \sim D_{tr} \\ g \sim \mathcal{S}}}
\left [\frac{-\psi \left ( \substack{ f^{M}_\mathcal{E}(x, \theta), \\ f^{M}_\mathcal{E}(g(x), \theta)} \right )}{\sum\limits_{\substack{x',y' \sim D_{tr} \\ y' \neq y}}
\psi \left ( \substack{f_{\mathcal{E}}^M(x, \theta),\\ f_{\mathcal{E}}^M(g(x'), \theta)} \right )
} \right ]
\\
&= \mathop{argmin}_{\theta} \mathop{\mathbb{E}}_{\substack{x, y \sim D_{tr} \\g \sim \mathcal{S}}}
\left [ -log \left (
\frac{\psi \left (  \substack{f^{M}_\mathcal{E}(x, \theta), \\ f^{M}_\mathcal{E}(g(x), \theta)} \right )}{\sum\limits_{\substack{x',y' \sim D_{tr} \\ y' \neq y}}
\psi \left ( \substack{f_{\mathcal{E}}^M(x, \theta),\\ f_{\mathcal{E}}^M(g(x'), \theta)} \right )
} \right ) \right ]
\end{split}
\end{equation}


We set the similarity metric, $\psi$, to be the same as our contrastive loss: $\psi(\cdot) = exp(\psi^{(cos)}(\cdot))$.

\begin{equation}
\label{eqn:proof}
\theta^* = \mathop{argmin}_{\theta} \mathop{\mathbb{E}}_{x, y \sim D_{train}} 
\left [
\mathcal{L}_{NCE}(x,y,\theta;\mathcal{S}) \right ]
\end{equation}

Here, we showed that the vanilla contrastive loss function, Equation~\ref{eqn:nce}, serves as a surrogate objective for optimizing our desired invariance objective, Equation~\ref{eqn:obj}. By incorporating contrastive learning alongside the maximum likelihood objective in Equation~\ref{eqn:ilo}, \ilo effectively reveals the underlying invariances in the pruned model.

\begin{table}
\centering
\begin{tabular}[width=0.5\textwidth]{| c | c | c c |} 
 \hline
Dataset & $g(\cdot)$ & \mlpv & \ourmlpv \\
\hline
\multirow{4}{*}{CIFAR10}
 & resize. & 44.096 $\pm$ 0.434 & {\bf 97.349 $\pm$ 4.590} \\
 & horiz. & 80.485 $\pm$ 0.504 & {\bf 99.413 $\pm$ 1.016} \\
 & color. & 56.075 $\pm$ 0.433 & {\bf 98.233 $\pm$ 3.060} \\
 & graysc. & 81.932 $\pm$ 0.233 & {\bf 99.077 $\pm$ 1.598} \\
\hline
\multirow{4}{*}{CIFAR100}
 & resize. & 32.990 $\pm$ 1.065 & {\bf 39.936 $\pm$ 2.786} \\
 & horiz. & 70.793 $\pm$ 0.677 & {\bf 77.935 $\pm$ 1.464} \\
 & color. & 31.704 $\pm$ 0.560 & {\bf 51.397 $\pm$ 2.709} \\
 & graysc. & 71.245 $\pm$ 0.467 & {\bf 76.476 $\pm$ 1.245} \\
\hline
\multirow{4}{*}{SVHN}  
 & resize. & 36.708 $\pm$ 2.033 & {\bf 77.440 $\pm$ 0.627} \\
 & horiz. & 71.400 $\pm$ 1.651 & {\bf 95.082 $\pm$ 0.166} \\
 & color. & 61.341 $\pm$ 0.946 & {\bf 91.097 $\pm$ 0.395} \\
 & graysc. & 90.344 $\pm$ 0.233 & {\bf 99.259 $\pm$ 0.073} \\
\hline
\end{tabular}
\caption{Comparing the consistency metric (\%) of the untrained supernetwork, \mlpv and \mlpt, against \our's pruned subnetwork under different invariant transforms, $g(\cdot)$. \our preserves invariances better.}
\label{tab:con2} 
\end{table}

\section{Additional Discussion on Lazy Training}

The lazy training regime~\citep{chizat2019lazy, tzen2020mean} is a phenomenon when loss rapidly decreases, while weight values stay relatively constant. This phenomenon occurs on large over-parameterized neural networks~\citep{chizat2019lazy}. Because the weight values stay relatively constant, the magnitude ordering between weights also changes very little. Therefore, network pruning struggles to preserve such loss decreases in the lazy training regime~\citep{liu2023sparsity}. 

Because weights with very small magnitude have minimal effect on the output logits, pruning said weights will not drastically hurt performance. Thus, if the pruning framework can separate very small magnitude weights from normal weights prior to the lazy training regime, we can preserve loss decreases in the lazy training regime. The \pis setting accomplishes this by initializing all weights to be very small so that only important weights will learn large magnitudes. This guarantees that a large percentage of weights will have small magnitudes throughout training, while important larger magnitude weights will emerge over the course of training.

%% file: sec-supp-exp.tex
\begin{figure}
\begin{tabular}{cc}
\subfloat[CIFAR10 ($\kappa=1.0$)]{\includegraphics[width = 1.5in]{figures/CIFAR10_weights_mlp.png}} &
\subfloat[CIFAR10 ($\kappa=0.125$)]{\includegraphics[width = 1.5in]{figures/CIFAR10_weights_our.png}} \\
\subfloat[CIFAR100 ($\kappa=1.0$)]{\includegraphics[width = 1.5in]{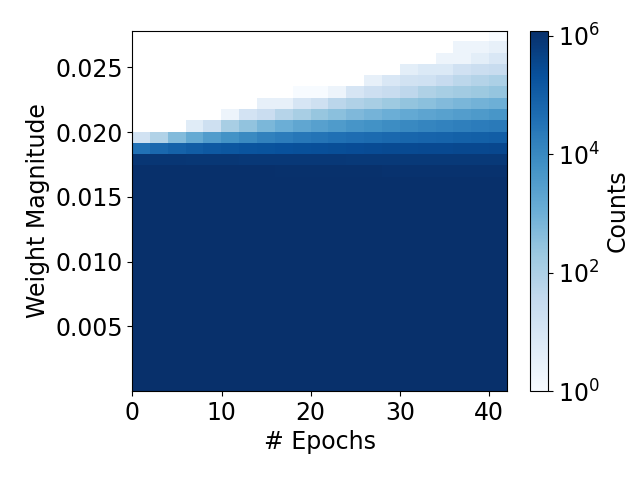}} &
\subfloat[CIFAR100 ($\kappa=0.125$)]{\includegraphics[width = 1.5in]{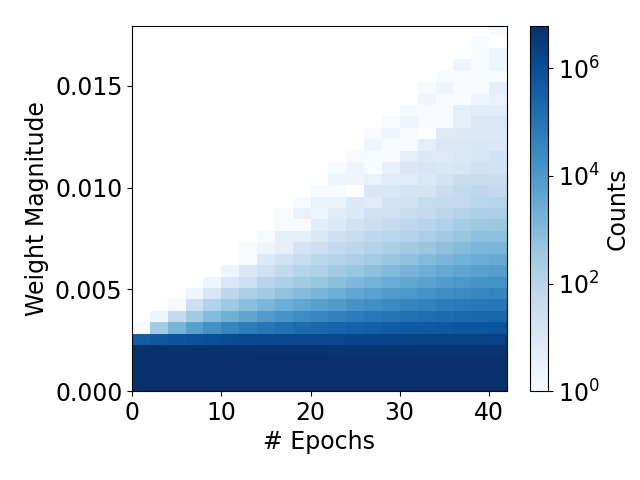}} \\
\subfloat[SVHN ($\kappa=1.0$)]{\includegraphics[width = 1.5in]{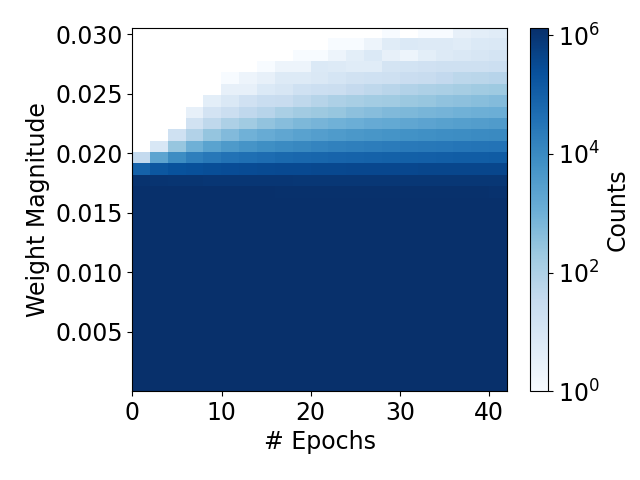}} &
\subfloat[SVHN ($\kappa=0.125$)]{\includegraphics[width = 1.5in]{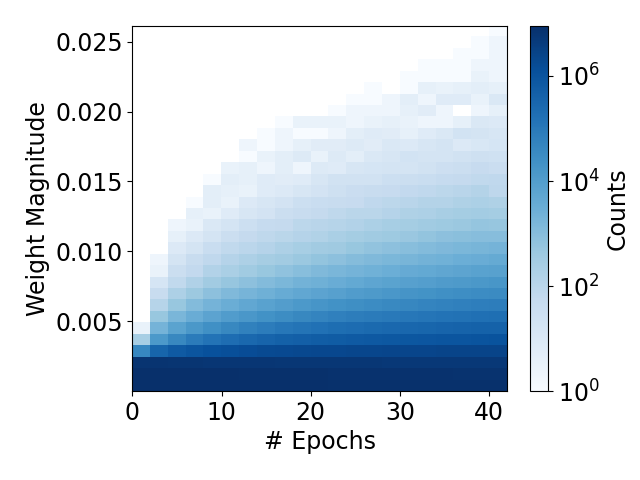}} 
\end{tabular}
\caption{Histogram of weight magnitudes, $|\theta_M^{(t)}|$, plotted over each epoch under different $\kappa$ initializations settings. $\kappa=1.0$ means normal initialization. Results shown for \mlpv on the CIFAR10, CIFAR100, and SVHN datasets.}
\label{fig:pis}
\end{figure}

\begin{figure}
\begin{tabular}{cc}
\subfloat[\mlpv CIFAR10]{\includegraphics[width = 1.5in]{figures/CIFAR10_img_mlp.png}} &
\subfloat[\ourmlpv CIFAR10]{\includegraphics[width = 1.5in]{figures/CIFAR10_img_our.png}} \\
\subfloat[\mlpv CIFAR100]{\includegraphics[width = 1.5in]{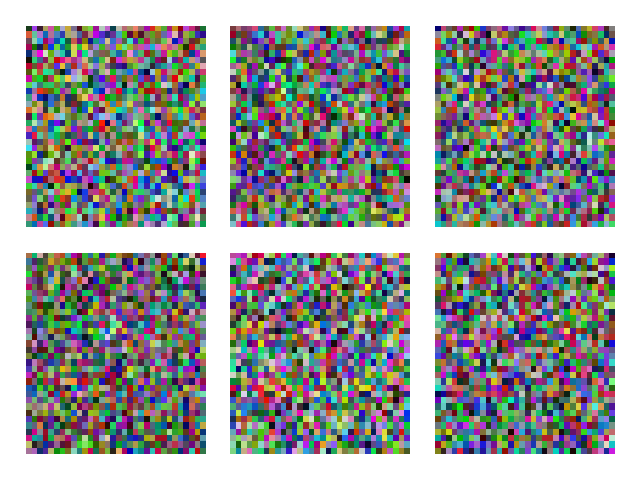}} &
\subfloat[\ourmlpv CIFAR100]{\includegraphics[width = 1.5in]{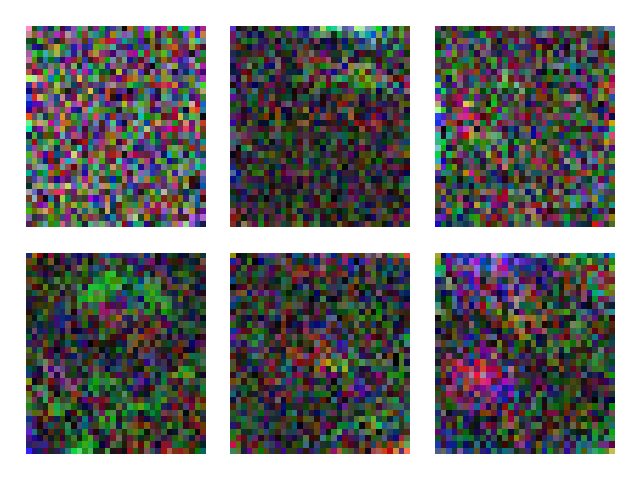}} \\
\subfloat[\mlpv SVHN]{\includegraphics[width = 1.5in]{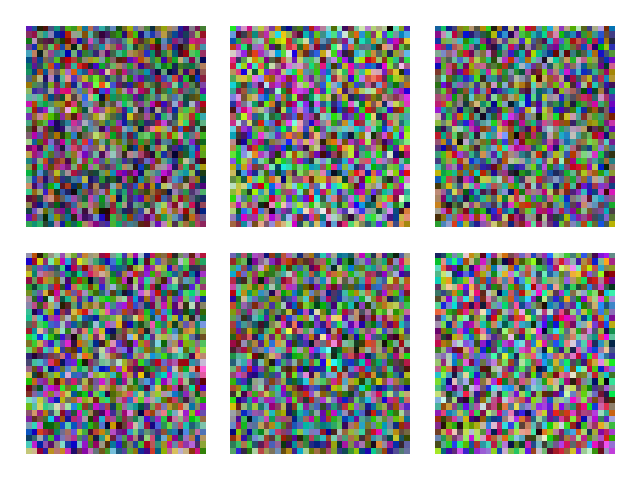}} &
\subfloat[\ourmlpv SVHN]{\includegraphics[width = 1.5in]{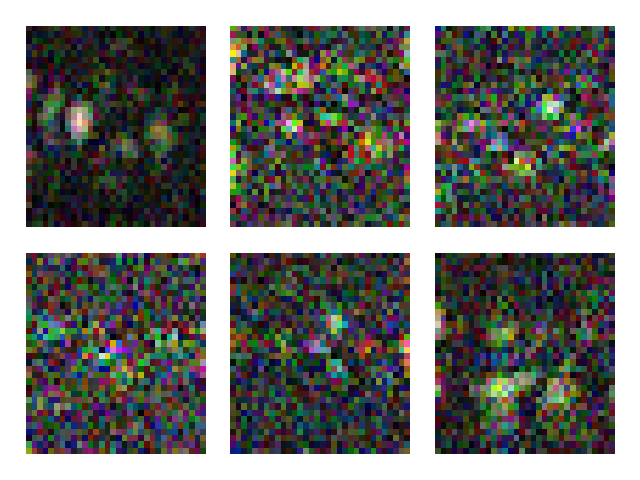}} \\
\end{tabular}
\caption{Visualization of weight magnitudes, $|\theta_M^{(T)}|$, trained with different policies. The models were trained on CIFAR10, CIFAR100, and SVHN. The magnitude of each RGB pixel for 6 output logits are plotted.}
\label{fig:vis}
\end{figure}

\section{Additional Experiments}

\subsection{Effects of Proactive Initialization: Full Results}

We provide weight histograms on CIFAR100 and SVHN in Figure~\ref{fig:pis}. As shown, the trends reported in the main text holds on other datasets.

\subsection{On Weight Visualization: Full Results}

We provide weight visualizations on CIFAR100 and SVHN in Figure~\ref{fig:vis}. As shown, the trends reported in the main text holds on other datasets.

\subsection{On Consistency: Full Results}

We provide consistency experiments on CIFAR100 and SVHN in Table~\ref{tab:con2}. As shown, the trends reported in the main text holds on other datasets.

%% file: sec-supp-implementation.tex
\section{Implementation Details}

\subsection{Dataset Details}

We considered the following {\it computer vision} datasets: CIFAR10, CIFAR100~\citep{krizhevsky2009learning}, and SVHN~\citep{netzer2011reading}. CIFAR10 and CIFAR100 are multi-domain image classification datasets. SVHN is a street sign digit classification dataset. Input images are $32\times32$ color images. We split the train set by 80/20 for training and validation. We test on the test set provided separately. We reported dataset statistics in Table~\ref{tab:data-cv}.

We considered 40 tabular datasets from OpenML~\citep{gijsbers2019open}, UCI~\citep{asuncion2007uci}, and Kaggle, following the \mlpc benchmark~\citep{kadra2021well}. These tabular datasets cover a variety of domains, data types, and class imbalances. We used a 60/20/20 train validation test split, and reported dataset statistics in Table~\ref{tab:data-tab}. We use a random seed of 11 for the data split, following prior work~\citep{kadra2021well}.

\subsection{Hyperparameter Settings}

All experiments were run 3 times from scratch starting with different random seeds. We report both the mean and standard deviation of all runs. All hyperparameters were chosen based on validation set results.

For all experiments, we used $\lambda=1$, which was chosen through a grid search over $\lambda \in \{0.25, 0.5, 1.0\}$. For all experiments, we used a batch size of 128. For pre-pruning training, we used SGD with Nesterov momentum and a learning rate of 0.001, following past works~\citep{blalock2020state}. For finetuning vision datasets, we used the same optimizer setup except with 16-bit operations except for batch normalization, following \betalasso~\citep{neyshabur2020towards}. For finetuning tabular datasets, we used AdamW~\citep{loshchilov2017decoupled}, a learning rate of 0.001s, decoupled weight decay, cosine annealing with restart, initial restart budget of 15 epochs, budget multiplier of 2, and snapshot ensembling~\citep{huang2017snapshot}, following prior works~\citep{kadra2021well,zimmer2021auto}. It is important to note we did not tune the dataset and training hyperparameters for each tabular dataset individually like \mlpc~\citep{kadra2021well}, rather taking the most effective setting on average.

For tabular datasets, we tuned the compression ratio over the following range of values: $r \in \{2,4,8\}$ and the \pis multiplier over the following range of values: $\kappa \in \{0.25, 0.125, 0.0625\}$. on a subset of 4 tabular datasets. We found that $r=8$ and $\kappa=0.25$ performs the most consistently and used this setting for all runs of \our in the main paper. It is important to note we did not tune hyperparameters for \our on each individual tabular dataset like \mlpc~\citep{kadra2021well}, making \our a much more efficient model than \mlpc. For the tabular baselines~\citep{chen2016xgboost, arik2021tabnet, kadra2021well}, we used the same hyperparameter tuning setup as the MLP+C benchmark~\citep{kadra2021well}.

For vision datasets, we tuned the compression ratio over the following range of values: $r \in \{2,4,8,16\}$ on each individual dataset for all network pruning models except \betalasso\footnote{This is because \betalasso does not accept a chosen compression ratio as a hyperparameter.}. For $\beta$-lasso~\citep{neyshabur2020towards}, we tuned the hyperparameters over the range $\beta=\{50\}$ and L1 regularization in $l1 \in \{10^{-6}, 2 \times 10^{-6}, 5 \times 10^{-6}, 10^{-5}, 2 \times 10^{-5}\}$ on each individual dataset as done in the original paper. It is important to note that although we tuned both hyperparameters for both \our and baselines on each individual datasets, our main and ablation table rankings stay consistent had we chosen a single setting for all datasets, as shown in the detailed pruning experiments in the main paper.

\subsection{Supernetwork Architecture}
\label{sec:arc}
\mlpv is a deep MLP that contains a CNN subnetwork. Given a scaling factor, $\alpha$, the CNN architecture consists of 3x3 convolutional layers with the following (out channels, stride) settings: [$(\alpha, 1)$, $(2\alpha, 2)$, $(2\alpha, 1)$, $(4\alpha, 2)$, $(4\alpha, 1)$, $(8\alpha, 2)$, $(8\alpha, 1)$, $(16\alpha, 2)$] followed by a hidden layer of dimension $64\alpha$. It is worth noting that our CNN does not include maxpooling layers for fair comparison with the learned architectures, following the same setup as \betalasso~\citep{neyshabur2020towards}. To form the MLP Network, we ensured the CNN network structure exists as a subnetwork within the MLP supernetwork by setting the hidden layer sizes to: $[\alpha s^2, \frac{\alpha s^2}{2} , \frac{\alpha s^2}{2}, \frac{\alpha s^2}{4}, \frac{\alpha s^2}{4}, \frac{\alpha s^2}{8}, \frac{\alpha s^2}{8}, \frac{\alpha s^2}{16}, 64\alpha]$. This architecture was also introduced in $\beta$-Lasso~\citep{neyshabur2020towards}. All layers are preceded by batch normalization and ReLU activation. We chose $\alpha=8$ such that our supernetwork can fit onto an Nvidia RTX 3070 GPU.

CNN is the corresponding CNN subnetwork with (out channels, stride) settings: [$(\alpha, 1)$, $(2\alpha, 2)$, $(2\alpha, 1)$, $(4\alpha, 2)$, $(4\alpha, 1)$, $(8\alpha, 2)$, $(8\alpha, 1)$, $(16\alpha, 2)$], derived from prior works~\citep{neyshabur2020towards}. Again, we chose $\alpha=8$ to be consistent with \mlpv.

\resnet~\citep{he2016deep} is the standard \resnet-18 model used in past benchmarks~\citep{blalock2020state}. Resnet differs from CNN in its inclusion of max-pooling layers and residual connections.

\mlpt is a 9-layer MLP with hidden dimension 512, batch normalization, and ReLU activation. We did not use dropout or skip connections as it was found to be ineffective on most tabular datasets in MLP+C~\citep{kadra2021well}.

\subsection{Pruning Implementation Details}

Following Shrinkbench~\citep{blalock2020state}, we use magnitude-based pruning only on the encoder, $f_{\mathcal{E}}$, keeping all weights in the decoder, $f_{\mathcal{D}}$. This is done to prevent pruning a cutset in the decoder architecture, so that all class logits receive input signal. To optimize the performance, we apply magnitude-based pruning globally, instead of layer-wise.

\subsection{Hardware}

All experiments were conducted on an Nvidia V100 GPU and an AMD EPYC 7402 CPU. The duration of the tabular experiments varied, ranging from a few minutes up to half a day, depending on the specific dataset-model pair and the training phase (pre-pruning training or finetuning). For the vision experiments, a single setting on a single dataset-model pair required a few hours for both pre-pruning training and finetuning.

\begin{table*}
\begin{center}
\begin{tabular}[width=0.5\textwidth]{|c | c c c | c |} 
 \hline
Dataset & \# Train Instances & \# Valid Instances & \# Test Instances & Number of Classes \\
\hline
CIFAR10 & 40000&10000&10000 & 10 \\
CIFAR100 & 40000&10000&10000 & 100 \\
SVHN & 58606&14651&26032 & 10 \\
\hline
\end{tabular}
\caption{Statistics on computer vision datasets.}
\label{tab:data-cv} 
\end{center}
\end{table*}

\begin{table*}
\begin{center}
\begin{tabular}[width=0.5\textwidth]{|c | c c c | c c c | c |} 
 \hline
\multirow{2}{*}{Dataset} & \# Train & \# Valid & \# Test & \multirow{2}{*}{\# Feats.} & Majority & Minority & \multirow{2}{*}{OpenML ID}\\
& Inst. & Inst. & Inst. & & Class \% & Class \%  & \\
\hline
Anneal&538 & 179 & 179&39&76.17&0.89&233090 \\
Kr-vs-kp&1917 & 639 & 639&37&52.22&47.78&233091 \\
Arrhythmia&271 & 90 & 90&280&54.20&0.44&233092 \\
Mfeat-factors&1200 & 400 & 400&217&10.00&10.00&233093 \\
Credit-g&600 & 200 & 200&21&70.00&30.00&233088 \\
Vehicle&507 & 169 & 169&19&25.77&23.52&233094 \\
Kc1&1265 & 421 & 421&22&84.54&15.46&233096 \\
Adult&29305 & 9768 & 9768&15&76.07&23.93&233099 \\
Walking-activity&89599 & 29866 & 29866&5&14.73&0.61&233102 \\
Phoneme&3242 & 1080 & 1080&6&70.65&29.35&233103 \\
Skin-segmentation&147034 & 49011 & 49011&4&79.25&20.75&233104 \\
Ldpa&98916 & 32972 & 32972&8&33.05&0.84&233106 \\
Nomao&20679 & 6893 & 6893&119&71.44&28.56&233107 \\
Cnae-9&648 & 216 & 216&857&11.11&11.11&233108 \\
Blood-transfusion&448 & 149 & 149&5&76.20&23.80&233109 \\
Bank-marketing&27126 & 9042 & 9042&17&88.30&11.70&233110 \\
Connect-4&40534 & 13511 & 13511&43&65.83&9.55&233112 \\
Shuttle&34800 & 11600 & 11600&10&78.60&0.02&233113 \\
Higgs&58830 & 19610 & 19610&29&52.86&47.14&233114 \\
Australian&414 & 138 & 138&15&55.51&44.49&233115 \\
Car&1036 & 345 & 345&7&70.02&3.76&233116 \\
Segment&1386 & 462 & 462&20&14.29&14.29&233117 \\
Fashion-MNIST&42000 & 14000 & 14000&785&10.00&10.00&233118 \\
Jungle-Chess-2pcs&26891 & 8963 & 8963&7&51.46&9.67&233119 \\
Numerai28.6&57792 & 19264 & 19264&22&50.52&49.48&233120 \\
Devnagari-Script&55200 & 18400 & 18400&1025&2.17&2.17&233121 \\
Helena&39117 & 13039 & 13039&28&6.14&0.17&233122 \\
Jannis&50239 & 16746 & 16746&55&46.01&2.01&233123 \\
Volkert&34986 & 11662 & 11662&181&21.96&2.33&233124 \\
MiniBooNE&78038 & 26012 & 26012&51&71.94&28.06&233126 \\
APSFailure&45600 & 15200 & 15200&171&98.19&1.81&233130 \\
Christine&3250 & 1083 & 1083&1637&50.00&50.00&233131 \\
Dilbert&6000 & 2000 & 2000&2001&20.49&19.13&233132 \\
Fabert&4942 & 1647 & 1647&801&23.39&6.09&233133 \\
Jasmine&1790 & 596 & 596&145&50.00&50.00&233134 \\
Sylvine&3074 & 1024 & 1024&21&50.00&50.00&233135 \\
Dionis&249712 & 83237 & 83237&61&0.59&0.21&233137 \\
Aloi&64800 & 21600 & 21600&129&0.10&0.10&233142 \\
C.C.FraudD&170884 & 56961 & 56961&31&99.83&0.17&233143 \\
Click Prediction&239689 & 79896 & 79896&12&83.21& 16.79&233146 \\
\hline
\end{tabular}
\caption{Statistics on tabular datasets. Note that the OpenML ID denotes the ID used to retrieve the dataset~\citep{gijsbers2019open}. Majority and Minority Class 
\% shows the class imbalance within each dataset. For fair evaluation, we report balanced accuracy in all tabular experiments. \# Feats. denotes the number of features in each dataset.}
\label{tab:data-tab} 
\end{center}
\end{table*}